\documentclass[runningheads]{llncs}
\usepackage{graphicx}
\usepackage{appendix}
\usepackage{amsfonts}
\usepackage{booktabs}
\usepackage{textcomp}
\usepackage{amssymb}
\usepackage{bbold}
\usepackage{multicol}
\usepackage{multirow}
\def\nbR{\ensuremath{\mathrm{I\! R}}}
\def\nbN{\ensuremath{\mathrm{I\! N}}}
\usepackage[
backend=biber,
style=alphabetic,
sorting=ynt
]{biblatex}
\begin{document}
\title{Influence branching for learning to solve mixed-integer programs online}
% \subtitle{Technical report for the MIP Workshop 2023 Computational Competition}
\author{Paul STRANG\inst{1, 2}, Zacharie ALES\inst{2,4}, Côme BISSUEL\inst{3}, Olivier JUAN\inst{3}, \\ Safia KEDAD-SIDHOUM\inst{4}, Emmanuel RACHELSON\inst{1}}
\authorrunning{P. Strang et al.}
\institute{ISAE-SUPAERO, Toulouse, France \and
ENSTA, Institut Polytechnique de Paris, France \and EDF R\&D, France
 \and CNAM Paris, CEDRIC, France}
\maketitle 
\begin{abstract}
On the occasion of the 20th Mixed Integer Program Workshop's computational competition, this work introduces a new approach for learning to solve MIPs online. Influence branching, a new graph-oriented variable selection strategy, is applied throughout the first iterations of the Branch \& Bound algorithm. This branching heuristic is optimized online with Thompson sampling, which ranks the best graph representations of MIP's structure according to computational speed up over SCIP. We achieve results comparable to state of the art online learning methods. Moreover, our results indicate that our method generalizes well to more general online frameworks, where variations in constraint matrix, constraint vector and objective coefficients can all occur and where more samples are available.

\end{abstract}
\section{Introduction}
    \label{section:intro}
    Mixed Integer Programming (MIP) is a subfield of combinatorial optimization, a discipline that aims at finding solutions to optimization problems with large but finite sets of feasible solutions. Research in the field has been in particular motivated by the countless industrial applications that can be derived in decision making and operations research. Mixed-integer program solvers developed over the last decades have relied on the Branch and Bound (B\&B) algorithm~\cite{wolsey2020integer} to efficiently explore the space of solutions while guaranteeing the optimality of the returned solution. Despite great discoveries in variable selection~\cite{achterberg_branching_2005}, node selection and cuts selection strategies, along with progress in the conception of primal heuristics and presolve methods~\cite{BestuzhevaEtal2021OO}, MIPs remain NP-hard problems for which computational load becomes intractable as the number of integer variables increases. Besides, the most efficient selection strategies are based on complex heuristics fine-tuned by experts on large MIP datasets to obtain the best average performance. In the context of real-world applications, in which similar instances with slightly varying inputs are solved on a regular basis, there is a huge incentive to reduce the solving time by learning efficient tailor-made heuristics.
    \newline The MIP Workshop 2023 Computational Competition \cite{bolusani2024mip} challenges participants to solve as fast as possible series of 50 MIP instances with slightly varying input, in an online fashion. In this work, we adopt machine learning's statistical point of view and think of each series $s \in \mathcal{S}$ as a set of 50 instances sampled from an unknown probability distribution $\mathcal{Q}_s$. Building on prior works~\cite{liberto_dash_2016,etheve_solving_2021,chmiela_online_2022}, we reformulate MIPcc23's reoptimization challenge as a multi-armed bandits problem, where the reward score $f_{s,i}(a)$ associated to instance $i \in \mathcal{I}_s$, is assumed to follow an unknown distribution $\mathcal{P}_{a, \, s}$ depending on $a \in \mathcal{A}$, the algorithm chosen to solve $i$. Then, $\sum_{i=1}^{50} f_{s,i}(a_i)$ is the sum of the rewards to minimize for each $s \in \mathcal{S}$, as to our designated action space,  $\mathcal{A}$, it consists in the set of hyperparameters couples used to parameterize our new variable selection strategy. SCIP with default parameters \cite{BestuzhevaEtal2021OO} is used as bedrock for our solution, while the implementation of our branching heuristic follows the \texttt{pyscipopt.Branchrule} API.  
    \newline This report is divided into 6 parts. In Section~\ref{section:background}, we motivate our approach by a short review of both the literature and the competition's guidelines. Section \ref{section:ibra} introduces influence branching, a new branching heuristic that leverages a graph representation of the current instance to select the most influential variable to branch on. Section~\ref{section:ibra} also highlights influence branching's speed up potential on the competition's instances, before underlining its sensitivity to hyperparameters tuning. This sensitivity justifies the need to learn online the best parameters for each series of instances. Section~\ref{section:bandits} briefly introduces the two online bandits algorithms considered to learn hyperparameters, and compare their average performances in terms of convergence. Section~\ref{section:results} provides an extensive presentation of computational results obtained on the competition's public instance series. Finally, Section~\ref{section:conclusion} concludes on the relevance of our approach for learning to solve mixed-integer programs online.

\section{Background}
    \label{section:background}
    
    Over the last decade, the field of machine learning has brought many contributions to the MIP literature \cite{huang_branch_2021, bengio_machine_2021}. Learning to cut \cite{huang_learning_2022,tang_reinforcement_2020}, learning to branch \cite{alvarez_supervised_2014, khalil_learning_2016, etheve_reinforcement_2020}, learning to select heuristics \cite{liberto_dash_2016, chmiela_online_2022} or learning to estimate solutions efficiently \cite{rachelson2010combining} are as many challenging tasks that the machine learning community has tried to address. However, the profile of the instances proposed by MIPcc23 along with specific requirements imposed by the competition's guidelines allow us to narrow the pool of promising approaches. The most binding requirement of the competition is undoubtedly its search for solutions that "work in practice". In fact, most contributions from the literature do not achieve better computational performances than state of the art solvers, and compare their solution to solvers with for example disabled presolve and cut generation plugins, in order to evaluate the speed up obtained by their sole input. In these conditions, works such as~\cite{gasse_exact_2019}, who achieved true state of the art performances on large MIP benchmarks by learning to imitate the strong branching strategy \cite{applegate1995finding}, appear promising. However, this achievement was made possible by the training of a large graph convolutionnal neural network on 100,000 samples drawn from 10,000 MIP instances. Such heavy computational work is not suited for our online setting, where every computation is the result of a trade off between learning a model and solving an instance. 
    \newline Turning to the competition's public series, apart from the \texttt{rhs 2} series, instances are rather hard to solve, leading to B\&B trees with at least 10,000 nodes and often much more. Thus, machine learning and in particular reinforcement learning approaches do not seem fitted for the challenge as they tend to scale poorly with dimension \cite{etheve_reinforcement_2020,scavuzzo_learning_2022}. Finally, given the high rate at which SCIP computes LP iterations, a lot of time would probably be lost in communication if we were to implement a python callback for every branching decision, i.e. at every node. 
    %% Maybe a couple of sentences to explain why we focus on the MIP representation and not on dynamic approach
    \newline For all these reasons, the MIPcc23 competition constitutes a challenging setup to apply branching strategy learning methods. In order to give the machine learning approach a chance to compete, the learning objective must be rationalized. Therefore, we propose to focus on learning graph representations leading to better branching decision near the root node, throughout the first iterations of the branch-and-bound algorithm. These graph representations, or influence models, and the maximal depth to apply our graph-based heuristic, are the two hyperparameters that will be optimized online with bandits algorithms.

\section{Influence branching}
    We introduce influence branching, a graph branching heuristic encoding part of the mixed-integer program structure into an influence graph which is then used for variable selection. Original works from~\cite{etheve_solving_2021} evidenced influence branching's great potential for tree size reduction, especially on hard instances. We provide an adapted version of this heuristic, which achieves better performances on the competition's instances. 
    \label{section:ibra}
    \subsection{Principle}
        Influence branching was first inspired by orbital branching \cite{ostrowski2011orbital}, a branching strategy that computes symmetry equivalent groups of variables to partition the search space into orbits. Similarly, influence branching as described by~\cite{etheve_solving_2021}, performs a clustering on an influence graph, a graph representation of the MIP instance, and splits variables into influence clusters. These clusters are then used by the brand-and-bound algorithm for variable selection, as it successively branches on each cluster throughout first iterations.
        In the following, we consider a mixed-integer linear program defined such as:
        \[ P : 
            \left\{
                \begin{array}{ll}
                    \min c^Tx \\
                    b^-\leq Ax \leq b^+ \; ; \; x \in \nbN^{|\mathcal{I}|} \times \nbR^{n-|\mathcal{I}|} 
                \end{array}
            \right.
        \]
        with $A \in \nbR^{m\times n}$, $b^{-},\,b^{+} \in \nbR^m$, $c \in \nbR^n$, $n$ the total number of variables, $m$ the number of linear constraints and $\mathcal{I}$ the indexes of integer variables.
        \begin{definition}(Local influence)
            We define the local influence $w_{ij}^l$ exerted by variable $i$ on variable $j$ through constraint $l$. $w_{ij}^l$ can be any function of A, b, c, in particular, we say that $i$ has a  non-zero influence on $j$ through $l$ if $\mathbb{1}_{A_{li}\ne  0}\mathbb{1}_{A_{lj}\ne 0} \ne 0$. 
        \end{definition}
        \begin{definition}(Direct influence)
            We define the direct influence $w_{ij}$ exerted by variable $i$ on variable $j$ over $P$ as :
            \[ w_{ij} = \mathbb{1}_{i \ne j} \, \sum_{l=1}^m w_{ij}^l\]
        \end{definition}
        We can then derive a definition for influence graphs. 
        \begin{definition}(Influence graph)
            We call influence graph the directed graph $G = (V, E, W)$ where  $V = \{1,\, ... , \, n\}$, $E = V \times V $ and where $W \in \nbR^{n\times n}$ the $w_{ij}$ matrix satisfies the definition of direct influence.
        \end{definition}
        Near the root node, taking the best branching decision does not necessarily mean branching on variables that will lead to the best immediate dual gap reduction, but rather means branching on variables that will have the most impact over the other variables in terms of integrity constraints. Therefore, influential variables are understood as variables which, when branched on, drive other variables to take the value of one of their bounds. Several factors making a variable influential are identified: being involved in a large number of constraints, being associated in average to larger constraint matrix coefficients, being involved in tight constraints or having large associated objective function coefficient.
        \newline Influence graphs are designed to capture maximum information from MIP instances' structure. We propose to evaluate the performance of several models of influence on the competition's public series, using $A$, $b$, $c$ as well as information extracted from the current LP iteration to define local influence. Influence models description can be found in Table \ref{tab:influence_models}. Although vectors $b$ and $c$ do not appear in the definition of local influence, they are used for the normalization of matrix $A$ which will be detailed in the next section.
        \begin{table}
            \begin{multicols}{2}
                \begin{description}
                    \item \textbf{Count} \hspace{2 mm} $w_{ij}^l = \mathbb{1}_{A_{li}}\mathbb{1}_{A_{lj}}$
                    \item \textbf{Binary} \hspace{2 mm} $w_{ij}^l = \frac{\mathbb{1}_{A_{li}}\mathbb{1}_{A_{lj}}}{\sum_{k=1}^m \mathbb{1}_{A_{ki}}\mathbb{1}_{A_{kj}}}$
                    \item \textbf{Dual} \hspace{2 mm} $w_{ij}^l = \mathbb{1}_{A_{li}}\mathbb{1}_{A_{lj}}
                    |y_{l}^*|$
                    \item \textbf{Countdual} \hspace{2 mm} $w_{ij}^l = \mathbb{1}_{A_{li}}\mathbb{1}_{A_{lj}}
                    \mathbb{1}_{({y_{l}}^* \ne 0)}$
                    \item \textbf{Auxiliary} \hspace{2 mm} $w_{ij}^l = \mathbb{1}_{A_{li}}\mathbb{1}_{A_{lj}} s_i |A_{li}y_l|$
                    \item \textbf{Adversarial} \hspace{2 mm} $w_{ij}^l = \mathbb{1}_{A_{li}}\mathbb{1}_{A_{lj}} s_i |\frac{A_{li}}{A_{lj}}|\mathbb{1}_{({y_{l}}^* \ne 0)}$
                \end{description}
            \end{multicols}
            \caption{Proposed influence models, with $y^*$ the solution of the dual problem at the current node and $s_i$ the minimal distance to a bound for variable $i$ in the primal solution. $\mathbb{1}_{A_{li}\ne  0}$ is noted $\mathbb{1}_{A_{li}}$ to ease the notations.}
            \label{tab:influence_models}
        \end{table}  
        
    \subsection{Adaptation to MIPcc23}
        
        Contrary to \cite{etheve_solving_2021}, we do not perform a spectral clustering on the influence graph. Our implementation of the influence branching heuristic returns the variable within the graph with the maximal total influence :
        \begin{equation}
            w_i = \sqrt{1+c_i}\sum_{j \ne i } w_{ij}(g)    
        \end{equation}
        as long as the depth of the current node $d$ is inferior or equal to $k$, the maximum depth. Moreover, variables' total influence are weighted according to their associated objective function coefficient $c_i$. For nodes of depth $d > k$, influence branching is disabled and the resolution of the MIP instance is handed over to SCIP set with default parameters. Before building the influence graph, a normalization of vectors $A$, $b$ and $c$ is carried out in order to make the matrix $W$ invariant to problem rescaling.
        \begin{equation}
            \left\{
                \begin{array}{ll}
                    c \leftarrow c/\sigma(c) \; if \; \sigma(c) \ne 0 \\
                    \\
                    A_k \leftarrow  A_k/b_k \; if \; b_k \ne 0 \\ 
                    \\
                    A_k \leftarrow  A_k/\sigma(A_k) \; if \; b_k = 0 \\ 
                \end{array}
            \right.
        \end{equation}
        with vector $b \in \nbR^m$ defined as $b_k = \mathbb{1}_{|b^+_k|<\infty}b^+_k - \mathbb{1}_{|b^-_k|<\infty}b^-_k $ for $k \in [1,\,m]$.
        \\
        \newline On hard instances, influence branching can achieve impressive speed up of solving time compared to state of the art solvers. Table \ref{tab:IBRA potential} provides the influence model $g$ and the maximal depth $k$ obtaining the best performance for each instance of \texttt{obj series 2}. As Table~\ref{tab:IBRA potential} highlights, the speed up potential of influence branching is significant, with an average speed up of -0.38 on \texttt{obj series 2} instances compared with default SCIP performances. However, Table~\ref{tab:IBRA potential} also exposes influence branching's extreme sensitivity to hyperparameters setting, as no pair $(g,k)$ appears to perform consistently better on every instance of the series. 
        
        \begin{table}[]
            \centering
            \begin{tabular}{c@{\hspace{.4cm}}c@{\hspace{.4cm}}c@{\hspace{.4cm}}c@{\hspace{.4cm}}c@{\hspace{.4cm}}c} \toprule
            \multirow{2}{*}{\textbf{Instance}} & \textbf{Influence } & \textbf{Max } & \multirow{2}{*}{\textbf{Performance}} & \textbf{SCIP} & \textbf{Speed} \\ 
            & \textbf{model} & \textbf{depth} & & \textbf{default} & \textbf{up}\\ \midrule
            1 & binary & 5 & 0.64& 0.70 & -0.06 \\
            2 & adversarial & 5 & 0.53 & 1.01 & -0.48 \\
            3 & countdual & 5 & 0.49 & 0.60 & -0.11 \\
            4 & count & 4 & 0.48 & 0.76 & -0.28 \\
            5 & countdual & 2 & 0.70 & 1.03 & -0.33 \\
            6 & count & 4 & 0.26 & 0.44 & -0.18 \\
            7 & auxiliary & 3 & 0.42 & 0.53& -0.11 \\
            8 & countdual & 2 & 0.71 & 0.98 & -0.37 \\
            9 & countdual & 4 & 0.57 & 1.39 & -0.82 \\
            10 & auxiliary & 1 & 0.35 & 1.00 & -0.65 \\
            11 & binary & 3 & 0.67 & 1.53 & -0.86 \\
            12 & dual & 1 & 0.84 & 1.19 & -0.35 \\
            13 & count & 3 & 0.24 & 0.45 & -0.21 \\
            14 & binary & 1 & 0.40 & 0.64 & -0.24 \\
            15 & countdual & 3 & 0.80 & 1.21 & -0.41 \\
            16 & auxiliary & 1 & 0.74 & 1.51 & -0.77 \\
            17 & binary & 1 & 0.87 & 1.21& -0.34 \\
            18 & countdual & 3 & 0.68 & 0.76 & 0.08 \\
            19 & binary & 1 & 0.28 & 0.45 & -0.17 \\
            ... &  ... &  ... &  ... &...& ... \\
            50 & binary & 5 & 0.29 & 0.66& -0.34 \\ \midrule
            \textbf{Avg} &  &  & \textbf{0.56}& \textbf{0.94} & \textbf{-0.38} \\
            
            \bottomrule
            \hspace{5mm}
            \end{tabular}
            \caption{Speed up potential of Influence branching on \texttt{obj series 2}. The performance column corresponds to $f_{s,\, i} = reltime + gap \, at \, time \, limit + nofeas$, while speed up indicates the performance gain obtained by influence branching compared to SCIP set with default parameters. To illustrate, for instance 2 the pair leading to the best performance of influence branching is $(g = adversarial, \, k = 5)$.} 
            \label{tab:IBRA potential}
        \end{table}

        Learning which pair $(g, \, k)$ performs best for any instance of any series would require to shift to a reinforcement learning framework, which would be both computationally unaffordable in the context of the competition, and theoretically challenging as it would require to find an efficient representation of MIP instances. Therefore, we adopt an online bandits framework, as we try to learn which pair $(g, \, k)$ obtains the best performance in average on a whole series of instances. The sorted average performance obtained by each pair $(g, \, k)$ on \texttt{obj series 2} is provided in Table \ref{tab:IBRA stats}. The average speed up obtained by the best pairs are promising enough to envisage online learning. Equivalent tables for the other series can be found in Appendix \ref{appendix:A}.
        \begin{table}[]
            \centering
            \begin{tabular}{c@{\hspace{.4cm}}c@{\hspace{.4cm}}c@{\hspace{.4cm}}c@{\hspace{.4cm}}c} \toprule
            \textbf{Influence model} & \textbf{Max depth} &\textbf{Performance} & \textbf{Speed up} & \textbf{Rank}\\ 
            \midrule
            count & 5 & 0.857 & \textbf{-0.0862} & \textbf{1} \\
            base & 6 & 0.865 & -0.0783 &2\\
            countdual & 2 & 0.874 & -0.0691 & 3 \\
            base & 5 & 0.877 & -0.0657 & 4\\
            count & 4 & 0.882 & -0.0606 & 5\\
            ... & ... & ... & ... & ...\\ 
            adversarial & 3 & 0.953 & 0.0520 & 34\\
            adversarial & 2 & 0.973 & 0.0721 & 35\\
            auxiliary & 5 & 1.05  & 0.148 & 36\\ 
            \bottomrule
            \hspace{5mm}
            \end{tabular}
            \caption{Sorted average performance of influence branching on \texttt{obj series 2} for each pair $(g,\,k)$. The performance column corresponds to the mean of $f_{s,\, i} = reltime + gap \, at \, time \, limit + nofeas$ over $\mathcal{I}_s$, while speed up indicates the average performance gain obtained by influence branching compared to SCIP set with default parameters.}
            \label{tab:IBRA stats}
        \end{table}
        
\section{Online bandits}
    As outlined in Sections~\ref{section:intro} and \ref{section:ibra}, instances from series $s \in \mathcal{S}$ are assumed to be sampled from an abstract distribution $\mathcal{Q}_{s}$ on the space of MIP instances, and scores $\{f_{s,\, i}(a)\}_{i\in \mathcal I_s}$  with  $ a \in \mathcal{A} = \{(g,k) : g \in \mathcal{G},\, k \in [1, 6]\} $ are assumed to follow an unknown probability distribution $\mathcal{P}_{a,\, s}$. The optimization task can be reformulated as a multi-armed bandits problem on action space $\mathcal{A}$ where 
    \begin{equation}
    \label{eq:objective}
        \min_{a_i \in \mathcal{A}} \, \sum_{i=1}^{50} (1+0.1i)\, f_{s,\,i}(a_i)
    \end{equation}
    is the sum of reward to minimize. We note $a^*_s$ the optimal action for series $s$, and $a_0$ the action corresponding to SCIP with default parameter.
    
\label{section:bandits}
    \subsection{Action space}
        For each series, only 50 samples are available in total. In order to minimize~(\ref{eq:objective}), the means of $(\mathcal{P}_{a,\, s})_{a\in \mathcal{A}}$, noted $(\mu_{a,\, s})_{a\in \mathcal{A}}$, need to be estimated (or at least ranked) as efficiently as possible for the heuristic to select the action leading to the expected minimum reward. The more actions in the action space, the more samples are needed to guarantee the convergence of the bandits algorithm towards optimal actions. Moreover, Table \ref{tab:IBRA stats} showed that the spreads between $(\mu_{a,\, s})_{a\in \mathcal{A}}$ are rather small, comprised between $0.01$ and $0.2$, in front of  standard deviations of $(\mathcal{P}_{a,\, s})_{a\in \mathcal{A}}$, noted $(\sigma_{a,\, s})_{a\in \mathcal{A}}$, that were measured around $0.1-0.3$ across public series. 
        \newline In order to mitigate identification issues, five actions among the best performing pairs $(g,\,k)$ across the competition's public series are selected to build action set $\mathcal{A}$:
        \begin{equation}
            \mathcal{A} = \{(count,\, 1), \, (count,\, 5), \, (countdual,\, 2), \,
                            (binary,\, 3), \, (dual,\, 3) \}
        \end{equation}
    \subsection{Algorithms}
        To find the best exploration-exploitation tradeoff, two bandits algorithms, Thompson sampling \cite{russo_tutorial_2020} and UCB2 \cite{auer_finite-time_2002} are evaluated on public instances series. For Thompson sampling, we make the simplifying assumption that  $(\mathcal{P}_{a,\, s})_{a\in \mathcal{A}}$ are normal distributions with unknown means $(\mu_{a,\, s})_{a\in \mathcal{A}}$ and fixed standard deviation $\sigma_{a,\, s} = \sigma = 0.2 $, the approximate value measured across public series. Thompson sampling algorithm draws samples from prior distributions $\mathcal{N}(\hat{\mu_{a}}, \, \hat{\sigma_{a}})$ corresponding to each action and selects the action associated with the minimum sampled value. Then, it collects a reward $r$, which corresponds to the performance obtained by the chosen algorithm on the instance, and performs a bayesian update of $\hat{\mu_{a}}$ and $\hat{\sigma_{a}}$. UCB2 on the other hand, does not require to make any additional assumption on $\mathcal{P}_{a,\, s}$. It selects the action minimizing the sum $\bar{x}_a + e_{a,\, r_a}$, with $\bar{x}_{a}$ the empirical mean of rewards associated with action $a$ and $e_{a,\, r_a}$ a measure of the exploration rate of action $a$. UCB2 has the advantage to be deterministic, its performance are reproducible while performances obtained by Thompson sampling must be averaged over a large number of runs.
    
    \subsection{Convergence}
        Convergence tests of online bandits algorithms are performed on \texttt{obj series 2} instances. In order to assess the robustness of the convergence, results are averaged over 10,000 runs. Instance series are shuffled before every run, so that instances are never solved in the same order. Also, it should be specified that computational results from Section \ref{section:ibra} are reused in this section and the next. In fact rewards are generated from the performance scores computed during the evaluation of influence branching: there is no further branch-and-bound tree being built at this point.
        %% Point out that no run was done, only data
        \begin{figure}
            \centering
            \includegraphics[width=8cm]{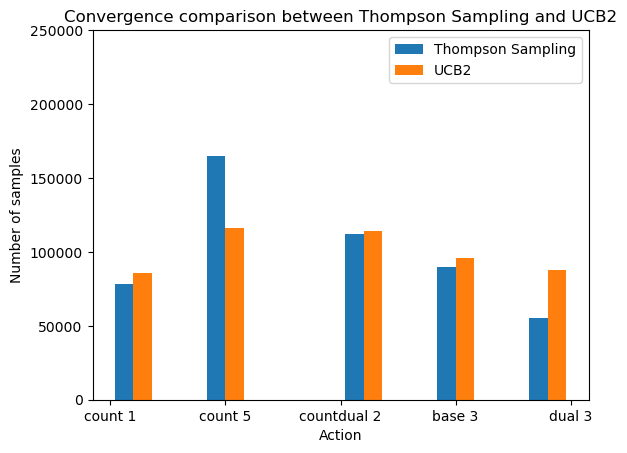}
            \caption{Convergence comparison between Thompson sampling and UCB2. Results of 10,000 on shuffled $obj \; series \; 2$ are aggregated.}
            \label{fig:histconvergence}
        \end{figure}
        
        UCB2 and Thompson sampling achieve very similar performance in terms of solving time across public series. However, as shown in Fig. \ref{fig:histconvergence}, Thompson sampling achieves better convergence towards optimal action $count \; 5$ than UCB2 over 50 iterations on \texttt{obj series 2}. This behaviour can be observed on every series.  The evaluation metric described in (\ref{eq:objective}) rewards submissions improving solving performance continuously over submission simply minimizing average performance, Thompson sampling is thus prefered over UCB2 for the implementation of our final solution.
        \newline Table~\ref{tab:regret} highlights the convergence performances of Thompson sampling across all public series averaged on 1000 runs. Once again, instance series are shuffled before every run. Convergence score is defined by:
        \begin{equation}
            CS = \frac{\sum_{i=1}^{50}\, \mu_{i,\,s}(a_i) - \mu_{i,\,s}(a_0)}{\sum_{i=1}^{50}\, \mu_{i,\,s}(a^*_s) - \mu_{i,\,s}(a_0)}
        \end{equation}
        with $\mu_{i,\,s}(a_i) - \mu_{i,\,s}(a_0)$ the expected speed up of action $a_i$ compared to SCIP and $\mu_{i,\,s}(a^*_s) - \mu_{i,\,s}(a_0)$ the speed up obtained by a theoretical oracle. For every series, a convergence score of at least 60 \% is reached. Consequently, on public series the average speed up achieved by our bandits algorithm is superior to half of the theoretical speed up obtained by an oracle performing the series' associated optimal action at every step. 
        \begin{table}[]
            \centering
            \begin{tabular}{c@{\hspace{.4cm}}c@{\hspace{.4cm}}c@{\hspace{.4cm}}c} \toprule
            \textbf{Series} & \textbf{Convergence score} \\ \midrule
             bnd series 1 & 72\% \\
             bnd series 2 &  65\%\\
             obj series 1 &  75\%\\
             obj series 2 &  66\%\\
             rhs series 1 &  64\%\\
             rhs series 2 &  72\%\\
             rhs obj series 1 & 74\% \\
            \bottomrule
            \hspace{5mm}
            \end{tabular}
            \caption{Convergence score of Thompson sampling on MIPcc23 public instances series.}
            \label{tab:regret}
        \end{table}

\section{MIPcc23 computational results}
    Table~\ref{tab:results} gathers measures of $f_{s,\,i}$ computed across every public series, including series with varying constraint matrix coefficients. Results are averaged over 2,000 runs with varying seed. The performance breakdown in terms of $reltime$, $dual \; gap$, $nofeas$ and $tree \, size$ can be found in Appendix~\ref{appendix:B}. Despite the fact that Thompson sampling converges towards optimal action, we don't observe better average performance on instances from the fifth batch. However, given the limited number of samples available, it is not sufficient to conclude that our solution is underfitting. This could result from the instance distribution of public series, for example from a concentration of harder instances in the last batches. Further discussion is provided in Section \ref{section:conclusion}.
    \begin{table}[]
        \centering
        \begin{tabular}{cc@{\hspace{4mm}}c@{\hspace{4mm}}c@{\hspace{4mm}}c@{\hspace{4mm}}c@{\hspace{4mm}}c} \toprule
         \textbf{Average $f_{s,\,i}$ scores}& \textbf{1-50} & \textbf{1-10}& \textbf{11-20}& \textbf{21-30}& \textbf{31-40}& \textbf{41-50} \\ \midrule
         \textbf{bnd series 1} & $0.992\pm 0.009$ & $1.036$ & $1.033$ & $0.947$ & $1.016$ & \textbf{0.930} \\
         \textbf{bnd series 2} &  $0.881\pm 0.020$  & 0.928 & \textbf{0.850} & 0.853 & 0.858 & 0.917\\
         \textbf{obj series 1} &  $0.895\pm 0.006$ &  \textbf{0.679} & 0.809 & 0.984 & 1.000 & 1.002\\
         \textbf{obj series 2} & $0.891\pm 0.022$& 0.847 & 1.018 & \textbf{0.825} & 0.859 & 0.910\\
         \textbf{rhs series 1} &  $0.875\pm 0.027 $& \textbf{0.810} & 0.865 & 0.839 & 0.906 & 0.954\\
         \textbf{rhs series 2} &  $1.004\pm 0.0001 $& 1.004 & 1.004 & 1.004 & 1.004 & 1.003\\
         \textbf{rhs obj series 1} & $1.015 \pm 0.006$ & 1.031 & 1.033 & 1.005 & \textbf{1.002} & 1.003\\ 
         \textbf{mat series 1} & $1.050 \pm  0.013 $ & \textbf{1.004}& 1.044 & 1.074 & 1.068 & 1.087\\ 
         \textbf{mat rhs bnd obj series 1} & $0.677 \pm 0.021  $& 0.768 & 0.707 & 0.660 & \textbf{0.538} & 0.714\\
        \bottomrule
        \hspace{5mm}
        \end{tabular}
        \caption{Average $f_{s,\, i}$ results. Instances are solved in the order of the competition dataset. Results are averaged over 2,000 runs, with varying seed. As a reminder, $f_{s,\, i}=  reltime + gap + nofeas$.}
        \label{tab:results}
    \end{table}
    \newline Table \ref{tab:speedup} highlights the speed up obtained by our solution over SCIP across public series. The "easy" instance series \texttt{rhs series 2} is the only series where no speed up is achieved. This does not come as a surprise, as \cite{etheve_solving_2021} showed that influence branching reduces the tree size of instances with large associated B\&B trees, while the size from \texttt{rhs series 2} instances' trees never exceeds a few dozen nodes. Our solution obtains good speed up performance on \texttt{bnd series 1}, \texttt{bnd series 2}, \texttt{obj series 1}, \texttt{obj series 2}, \texttt{rhs series 1} and \texttt{mat rhs bnd obj rhs series 1}, with an average score reduction located between -0.02 and -0.06. By comparison, \cite{chmiela_online_2022} achieves a 4\% speed up over SCIP only on instances taking more than 1000 seconds to solve, while training over 175 instances. This highlights the capacity of influence branching to leverage part of the MIP instance structure to perform efficient var iable selection near the root node. Turning to \texttt{rhs obj series 1} and \texttt{mat series 1}, the speed up obtained may turn out to be more significant than it appears, for the average score improvement corresponds to a reduction of the average dual gap at termination, and not to a reduction of solving time. Comparison with other candidates scores is necessary to assess our performance on these series.
    % This also stands for \texttt{mat series 1}, upon which limited speed up is also achieved.
    % Finally, $mat \, series \, 1$ also achieves limited performance speed up. % This results from a deliberate choice. Since this series is excluded from the pool of series used for final evaluation, we did not include in our action set $\mathcal{A}$ the pairs $(g,\, k)$ leading to the best performance speed up, in order to improve the convergence of Thompson sampling on the other public series.
    \begin{table}[]
        \centering
        \begin{tabular}{c@{\hspace{4mm}}c@{\hspace{8mm}}c} \toprule
         \textbf{Series}& \textbf{Average $f_{s,\, i}$} & \textbf{Speed up compared to SCIP}\\ \midrule
         \textbf{bnd series 1} & $0.992\pm 0.009$ & \boldmath$-0.031\pm 0.009$ \\
         \textbf{bnd series 2} &  $0.881\pm 0.020$  &  \boldmath$-0.037\pm 0.020$\\
         \textbf{obj series 1} &  $0.895\pm 0.006$ &  \boldmath$-0.022\pm 0.006$\\
         \textbf{obj series 2} & $0.891\pm 0.022$ & \boldmath$-0.052\pm 0.022$\\
         \textbf{rhs series 1} &  $0.875\pm 0.027 $ &  \boldmath$-0.048\pm 0.027 $\\
         \textbf{rhs series 2} &  $1.004\pm 0.0001 $ &  $0.001 \pm 0.0001 $\\
         \textbf{rhs obj series 1} & $1.015 \pm 0.006$ & $-0.005 \pm 0.006$\\ 
         \textbf{mat series 1} & $1.050 \pm  0.013 $ & $-0.009 \pm  0.013 $ \\ 
         \textbf{mat rhs bnd obj series 1} & $0.677 \pm 0.021 $ & \boldmath$-0.061 \pm 0.021 $\\
        \bottomrule
        \hspace{5mm}
        \end{tabular}
        \caption{Averaged speed up obtained across public series. Results are averaged over 2,000 runs, with varying seed.}
        \label{tab:speedup}
    \end{table}

    \label{section:results}

\section{Perspectives}
    \label{section:conclusion}
    Learning to solve MIP instances online is a challenging task. Adopting a bandits framework, this work proposes to learn online the optimal setting of influence branching among five preselected pairs $(g,\, k)$ of parameters. Since these pairs were selected according to their performance on public series, $s \in \mathcal{S}$, it is legitimate to wonder whether performances on hidden series, ${s' \in \mathcal{S}'}$, will be comparable. Two arguments can be made in favour of our approach.
    \newline First, as highlighted in Appendix \ref{appendix:A}, suboptimal actions also lead to significant average speed ups. Consequently, it is likely that for a hidden series sampled from an unknown distribution $\mathcal{Q}_{s'}$, one or several actions from $\mathcal{A}$ will lead to an average performance speed up. Second, even in the case when none of the 5 actions of our action set leads to an average performance speed up, this does not disqualify our approach for learning to solve MIPs online. In fact, owing to the limited number of samples available for each series of the competition, we were constrained to reduce the size of our action set to guarantee the convergence of Thompson sampling. However, in a more general framework where possibly several hundreds of instances sampled from the same distribution are solved online \cite{chmiela_online_2022}, larger action sets could be built while preserving Thompson sampling convergence properties. This would result in a better adaptability of our solution to instance series sampled from unknown distributions $(\mathcal{Q}_{s'})_{s' \in \mathcal{S'}}$ while preserving performances obtained on instance series sampled from $(\mathcal{Q}_{s})_{s \in \mathcal{S}}$. 
    % Furthermore (very general approach, advantage).

\printbibliography
\newpage
\appendix
\section{Influence branching average speed up potential on public series}
\label{appendix:A}
    \begin{table}[]
        \centering
        \begin{tabular}{c@{\hspace{.4cm}}c@{\hspace{.4cm}}c@{\hspace{.4cm}}cc} \toprule
        \textbf{Influence } & \textbf{Max } & \multirow{2}{*}{\textbf{Performance}} & \textbf{Speed} \\ 
         \textbf{model} & \textbf{depth} & & \textbf{up}\\ \midrule
        count & 1 & 0.980 & \textbf{-0.0424} \\
        binary & 1 & 0.981 & -0.0418 \\
        binary & 3 & 0.983 & -0.0392 \\
        count & 3 & 0.990 & -0.0324 \\
        dual & 1 & 0.991 & -0.0312 \\
        ... & ... & ... & ... \\ 
        dual & 6 & 1.031 & +0.0084 \\
        countdual & 4 & 1.032 & +0.0090 \\
        countdual & 6 & 1.041  & +0.0190 \\ 
        \bottomrule
        \hspace{5mm}
        \end{tabular}
        \caption{Sorted average performance of influence branching on \textbf{\texttt{bnd series 1}} for each pair $(g,\,k)$.}
    \end{table}
    \begin{table}[]
        \centering
        \begin{tabular}{c@{\hspace{.4cm}}c@{\hspace{.4cm}}c@{\hspace{.4cm}}cc} \toprule
        \textbf{Influence } & \textbf{Max } & \multirow{2}{*}{\textbf{Performance}} & \textbf{Speed} \\ 
         \textbf{model} & \textbf{depth} & & \textbf{up}\\ \midrule
        countdual & 2 & 0.845 & \textbf{-0.0733} \\
        countdual & 3 & 0.846 & -0.0721 \\
        countdual & 1 & 0.865 & -0.0527 \\
        dual & 1 & 0.869 & -0.0492 \\
        count & 1 & 0.870 & -0.0482 \\
        ... & ... & ... & ... \\ 
        dual & 5 & 0.983 & +0.0654 \\
        dual & 6 & 0.996 & +0.0774 \\
        dual & 4 & 0.996  & +0.0782 \\ 
        \bottomrule
        \hspace{5mm}
        \end{tabular}
        \caption{Sorted average performance of influence branching on \textbf{\texttt{bnd series 2}} for each pair $(g,\,k)$.}
    \end{table}
    \begin{table}[]
        \centering
        \begin{tabular}{c@{\hspace{.4cm}}c@{\hspace{.4cm}}c@{\hspace{.4cm}}cc} \toprule
        \textbf{Influence } & \textbf{Max } & \multirow{2}{*}{\textbf{Performance}} & \textbf{Speed} \\ 
         \textbf{model} & \textbf{depth} & & \textbf{up}\\ \midrule
        binary & 3 & 0.884 & \textbf{-0.0330} \\
        countdual & 5 & 0.885 & -0.0322 \\
        dual & 5 & 0.886 & -0.0311 \\
        dual & 4 & 0.886 & -0.0308 \\
        count & 3 & 0.888 & -0.0290 \\
        ... & ... & ... & ... \\ 
        countdual & 1 & 0.904 & -0.0098 \\
        count & 1 & 0.906 & -0.0081 \\
        binary & 1 & 0.907  & -0.0070 \\ 
        \bottomrule
        \hspace{5mm}
        \end{tabular}
        \caption{Sorted average performance of influence branching on \textbf{\texttt{obj series 1}} for each pair $(g,\,k)$.}
    \end{table}
    \begin{table}[]
        \centering
        \begin{tabular}{c@{\hspace{.4cm}}c@{\hspace{.4cm}}c@{\hspace{.4cm}}cc} \toprule
        \textbf{Influence } & \textbf{Max } & \multirow{2}{*}{\textbf{Performance}} & \textbf{Speed} \\ 
         \textbf{model} & \textbf{depth} & & \textbf{up}\\ \midrule
        count & 1 & 0.835 & \textbf{-0.0872} \\
        binary & 1 & 0.848 & -0.0746 \\
        countdual & 6 & 0.855 & -0.0676 \\
        dual & 5 & 0.858 & -0.0643 \\
        countdual & 3 & 0.863 & -0.0598 \\
        ... & ... & ... & ... \\ 
        binary & 5 & 0.914 & -0.0087 \\
        count & 3 & 0.920 & -0.00321 \\
        binary & 5 & 0.926  & +0.0031 \\ 
        \bottomrule
        \hspace{5mm}
        \end{tabular}
        \caption{Sorted average performance of influence branching on \textbf{\texttt{rhs series 1}} for each pair $(g,\,k)$.}
    \end{table}
    \begin{table}[]
        \centering
        \begin{tabular}{c@{\hspace{.4cm}}c@{\hspace{.4cm}}c@{\hspace{.4cm}}cc} \toprule
        \textbf{Influence } & \textbf{Max } & \multirow{2}{*}{\textbf{Performance}} & \textbf{Speed} \\ 
         \textbf{model} & \textbf{depth} & & \textbf{up}\\ \midrule
        count & 1 & 1.003 & +0.0008 \\
        countdual & 1 & 1.003 & +0.0009 \\
        binary & 1 & 1.003 & +0.0010 \\
        countdual & 2 & 1.004 & +0.0010 \\
        dual & 1 & 1.004 & +0.0011 \\
        ... & ... & ... & ... \\ 
        binary & 6 & 1.004 & +0.0016 \\
        count & 6 & 1.004 & +0.0016 \\
        binary & 5 & 1.004  & +0.0016 \\ 
        \bottomrule
        \hspace{5mm}
        \end{tabular}
        \caption{Sorted average performance of influence branching on \textbf{\texttt{rhs series 2}} for each pair $(g,\,k)$.}
    \end{table}
    \begin{table}[]
        \centering
        \begin{tabular}{c@{\hspace{.4cm}}c@{\hspace{.4cm}}c@{\hspace{.4cm}}cc} \toprule
        \textbf{Influence } & \textbf{Max } & \multirow{2}{*}{\textbf{Performance}} & \textbf{Speed} \\ 
         \textbf{model} & \textbf{depth} & & \textbf{up}\\ \midrule
        dual & 3 & 1.012 & \textbf{-0.0084} \\
        count & 8 & 1.013 & -0.0078 \\
        binary & 8 & 1.013 & -0.078 \\
        binary & 2 & 1.014 & -0.0066 \\
        count & 2 & 1.014 & -0.0066\\
        ... & ... & ... & ... \\ 
        binary & 4 & 1.027 & +0.0065\\
        countdual & 2 & 1.032 & +0.0011\\
        dual & 1 & 1.032 & +0.0012 \\ 
        \bottomrule
        \hspace{5mm}
        \end{tabular}
        \caption{Sorted average performance of influence branching on \textbf{\texttt{rhs obj series 1}} for each pair $(g,\,k)$.}
    \end{table}
    \begin{table}[]
        \centering
        \begin{tabular}{c@{\hspace{.4cm}}c@{\hspace{.4cm}}c@{\hspace{.4cm}}cc} \toprule
        \textbf{Influence } & \textbf{Max } & \multirow{2}{*}{\textbf{Performance}} & \textbf{Speed} \\ 
         \textbf{model} & \textbf{depth} & & \textbf{up}\\ \midrule
        count & 1 & 1.020 & \textbf{-0.0364} \\
        countdual & 1 & 1.030 & -0.0270 \\
        binary & 1 & 1.034 & -0.0237 \\
        count & 2 & 1.040 & -0.0170 \\
        binary & 2 & 1.041 & -0.0162 \\
        ... & ... & ... & ... \\ 
        dual & 5 & 1.110 & +0.0540 \\
        dual & 6 & 1.226 & +0.0656 \\
        countdual & 6 & 1.140  & +0.0831 \\ 
        \bottomrule
        \hspace{5mm}
        \end{tabular}
        \caption{Sorted average performance of influence branching on \textbf{\texttt{mat series 1}} for each pair $(g,\,k)$.}
    \end{table}
    \begin{table}[]
        \centering
        \begin{tabular}{c@{\hspace{.4cm}}c@{\hspace{.4cm}}c@{\hspace{.4cm}}cc} \toprule
        \textbf{Influence } & \textbf{Max } & \multirow{2}{*}{\textbf{Performance}} & \textbf{Speed} \\ 
         \textbf{model} & \textbf{depth} & & \textbf{up}\\ \midrule
        dual & 3 & 0.643 & \textbf{-0.0957} \\
        dual & 2 & 0.653 & -0.0852 \\
        count & 1 & 0.659 & -0.0796 \\
        count & 2 & 0.664 & -0.0745 \\
        countdual & 3 & 0.670 & -0.0679 \\
        ... & ... & ... & ... \\ 
        count & 6 & 0.726 & -0.0129 \\
        countdual & 6 & 0.730 & -0.0083 \\
        count & 5 & 0.745  & +0.0063 \\ 
        \bottomrule
        \hspace{5mm}
        \end{tabular}
        \caption{Sorted average performance of influence branching on \textbf{\texttt{mat rhs bnd obj series 1}} for each pair $(g,\,k)$.}
    \end{table}
\newpage
\section{Performance breakdown on public series}
\label{appendix:B}
    \begin{table}[]
        \centering
        \begin{tabular}{cc@{\hspace{4mm}}c@{\hspace{4mm}}c@{\hspace{4mm}}c@{\hspace{4mm}}c@{\hspace{4mm}}c} \toprule
         \textbf{Average $tree \, size$}& \textbf{1-50} & \textbf{1-10}& \textbf{11-20}& \textbf{21-30}& \textbf{31-40}& \textbf{41-50} \\ \midrule
         \textbf{bnd series 1} & $7449\pm 216$ & $7222$ & $6832$ & $9550$ & $7277$ & \textbf{6364} \\
         \textbf{bnd series 2} & $10346\pm 369$ & $10355$ & $10138$ & $10208$ & $10586$ & \textbf{10442} \\
         \textbf{obj series 1} &  $245255\pm 2825$ &  \textbf{181684} & 228534 & 290934 & 275192 & 249931\\
         \textbf{obj series 2} & $87956\pm 2077$& 99270 & 85366 & 88066 & \textbf{81301} & 85774\\
         \textbf{rhs series 1} &  $22143\pm 999 $& \textbf{17996} & 21930 & 25537 & 20679 & 24571\\
         \textbf{rhs series 2} &  $37\pm 1 $& 43 & 42 & 14 & 38 & 48\\
         \textbf{rhs obj series 1} & $587 \pm 40$ & 498 & 225 & 569 & 899 & 741\\ 
         \textbf{mat series 1} & $9084\pm  295 $ & 8902 & 9294 & 9384 & 8850 & 8986\\ 
         \textbf{mat rhs bnd obj series 1} & $3332 \pm 129  $& 3711 & 3632 & 3198 & \textbf{2053} & 4066\\
        \bottomrule
        \hspace{5mm}
        \end{tabular}
        \caption{$Tree\, size$ results. Instances are solved in the order of the competition dataset. Results are averaged over 2,000 runs, with varying seed. }
    \end{table}
    \begin{table}[]
        \centering
        \begin{tabular}{cc@{\hspace{4mm}}c@{\hspace{4mm}}c@{\hspace{4mm}}c@{\hspace{4mm}}c@{\hspace{4mm}}c} \toprule
         \textbf{Average $reltime$}& \textbf{1-50} & \textbf{1-10}& \textbf{11-20}& \textbf{21-30}& \textbf{31-40}& \textbf{41-50} \\ \midrule
         \textbf{bnd series 1} & $0.949\pm 0.009$ & $0.980$ & $0.992$ & $0.914$ & $0.965$ & \textbf{0.895} \\
         \textbf{bnd series 2} &  $0.842\pm 0.019$  & 0.884 & 0.817 & \textbf{0.811} & 0.824 & 0.879\\
         \textbf{obj series 1} &  $0.895\pm 0.006$ &  \textbf{0.676} & 0.815 & 0.984 & 1.000 & 1.000\\
         \textbf{obj series 2} & $0.802\pm 0.014$& 0.822 & 0.848 & 0.777 & \textbf{0.741} & 0.823\\
         \textbf{rhs series 1} &  $0.874\pm 0.029 $& \textbf{0.805} & 0.862 & 0.841 & 0.913 & 0.949\\
         \textbf{rhs series 2} &  $1.000\pm 0.000 $& 1.000 & 1.000 & 1.000 & 1.000 & 1.000\\
         \textbf{rhs obj series 1} & $0.998 \pm 0.001$ & \textbf{0.988} & 1.000 & 1.000 & 1.000 & 1.000\\ 
         \textbf{mat series 1} & $0.971 \pm  0.008 $ & \textbf{0.947}& 0.971 & 0.968 & 0.977 & 0.996\\ 
         \textbf{mat rhs bnd obj series 1} & $0.677 \pm 0.021  $& 0.768 & 0.707 & 0.660 & \textbf{0.538} & 0.714\\
        \bottomrule
        \hspace{5mm}
        \end{tabular}
        \caption{$reltime$ results. Instances are solved in the order of the competition dataset. Results are averaged over 2,000 runs, with varying seed. }
    \end{table}
    \begin{table}[]
        \centering
        \begin{tabular}{cc@{\hspace{4mm}}c@{\hspace{4mm}}c@{\hspace{4mm}}c@{\hspace{4mm}}c@{\hspace{4mm}}c} \toprule
         \textbf{Average $dual \, gap$}& \textbf{1-50} & \textbf{1-10}& \textbf{11-20}& \textbf{21-30}& \textbf{31-40}& \textbf{41-50} \\ \midrule
         \textbf{bnd series 1} & $0.045\pm 0.002$ & $0.057$ & $0.042$ & $0.039$ & $0.050$ & \textbf{0.038} \\
         \textbf{bnd series 2} &  $0.036\pm 0.004$  & 0.048 & 0.034 & 0.036 & \textbf{0.021} & 0.039\\
         \textbf{obj series 1} &  $0.001\pm 0.000$ &  \textbf{0.000} & 0.000 & 0.001 & 0.001 & 0.002\\
         \textbf{obj series 2} & $0.095\pm 0.013$& 0.029 & 0.175 & \textbf{0.046} & 0.129& 0.097\\
         \textbf{rhs series 1} &  $0.001\pm 0.000 $& \textbf{0.000} & 0.000 & 0.000 & 0.001 & 0.001\\
         \textbf{rhs series 2} &  $0.003\pm 0.0001 $& 0.003 & 0.003 & 0.003 & 0.003 & 0.003\\
         \textbf{rhs obj series 1} & $0.017 \pm 0.006$ & 0.039 & 0.038 & 0.005 & \textbf{0.002} & 0.003\\ 
         \textbf{mat series 1} & $0.075 \pm  0.007 $ & \textbf{0.053}& 0.071 & 0.073 & 0.084 & 0.094\\ 
         \textbf{mat rhs bnd obj series 1} & $0.006 \pm 0.003  $& 0.020 & 0.001 & \textbf{0.001} & 0.002 & 0.004\\
        \bottomrule
        \hspace{5mm}
        \end{tabular}
        \caption{$dual \; gap$ results. Instances are solved in the order of the competition dataset. Results are averaged over 2,000 runs, with varying seed. }
    \end{table}
    \begin{table}[]
        \centering
        \begin{tabular}{cc@{\hspace{4mm}}c@{\hspace{4mm}}c@{\hspace{4mm}}c@{\hspace{4mm}}c@{\hspace{4mm}}c} \toprule
         \textbf{Average $nofeas$}& \textbf{1-50} & \textbf{1-10}& \textbf{11-20}& \textbf{21-30}& \textbf{31-40}& \textbf{41-50} \\ \midrule
         \textbf{bnd series 1} & $0.000\pm 0.000$ & $0.000$ & $0.000$ & $0.000$ & $0.000$ & 0.000 \\
         \textbf{bnd series 2} &  $0.000\pm 0.000$  & 0.000 & 0.000 & 0.000 & 0.000 & 0.000\\
         \textbf{obj series 1} &  $0.000\pm 0.000$ &  0.000 & 0.000 & 0.000 & 0.000 & 1.002\\
         \textbf{obj series 2} & $0.000\pm 0.000$& 0.000 & 0.000 & 0.000 & 0.000 & 0.000\\
         \textbf{rhs series 1} &  $0.000\pm 0.000 $& 0.000 & 0.000 & 0.000 & 0.000 & 0.000\\
         \textbf{rhs series 2} &  $0.000\pm 0.000 $& 0.000 & 0.000 & 0.000 & 0.000 & 0.000\\
         \textbf{rhs obj series 1} & $0.000 \pm 0.000$ & 0.000 & 0.000 & 0.000 & 0.000 & 0.000\\ 
         \textbf{mat series 1} & $0.000 \pm  0.000 $ & 0.000& 0.000 & 0.000 & 0.000 & 0.000\\ 
         \textbf{mat rhs bnd obj series 1} & $0.000 \pm 0.000  $& 0.000 & 0.000 & 0.000 & 0.000 & 0.000\\
        \bottomrule
        \hspace{5mm}
        \end{tabular}
        \caption{$nofeas$ results. Instances are solved in the order of the competition dataset. Results are averaged over 2,000 runs, with varying seed. }
    \end{table}

\end{document}